# MASTER: Multimodal Segmentation with Text Prompts

Fuyang Liu[1,2], Shun Lu[1,2], Jilin Mei[1], and Yu Hu[1,2]

[1] Institute of Computing Technology, Chinese Academy of Sciences, Beijing, 100190, China
[2] University of Chinese Academy of Sciences, Beijing, 101408, China
`{liufuyang2023z,lushun19s, meijilin, huyu}@ict.ac.cn`

**Abstract.** RGB-Thermal fusion is a potential solution for various weather and light conditions in challenging scenarios. However, plenty of studies focus on designing complex modules to fuse different modalities. With the widespread application of large language models (LLMs), valuable information can be more effectively extracted from natural language. Therefore, we aim to leverage the advantages of large language models to design a structurally simple and highly adaptable multimodal fusion model architecture. We proposed **M**ultimod**A**l **S**egmentation with **TE**xt P**R**ompts (MASTER) architecture, which integrates LLM into the fusion of RGB-Thermal multimodal data and allows complex query text to participate in the fusion process. Our model utilizes a dual-path structure to extract information from different modalities of images. Additionally, we employ LLM as the core module for multimodal fusion, enabling the model to generate learnable codebook tokens from RGB, thermal images, and textual information. A lightweight image decoder is used to obtain semantic segmentation results. The proposed MASTER performs exceptionally well in benchmark tests across various automated driving scenarios, yielding promising results.

## 1 Introduction

Reliable understanding of the scene under varied scenarios is a challenging task in autonomous driving. A multimodal fusion strategy like RGB-Thermal fusion [1] is a feasible solution to tackle complex weather and light conditions. For example, the RGB camera is capable of capturing abundant texture and color information with good visibility but weak with inadequate light. Thermal cameras are robust in different light conditions but can only capture limited details. Therefore, it is reasonable to combine RGB and thermal channels as complementarities to improve the perception of autonomous driving. The key challenge is how to fuse information from different modalities. Most of the existing approaches [1-5] rely on concatenating or pooling across different channels for fusion. However, such methods are typically pyramidal, requiring fusion across multiple scales, which complicates the model structure. Structures based on previous fusion strategies are often asymmetric [6, 7], meaning that the processing strategies for different types of data are inflexible.

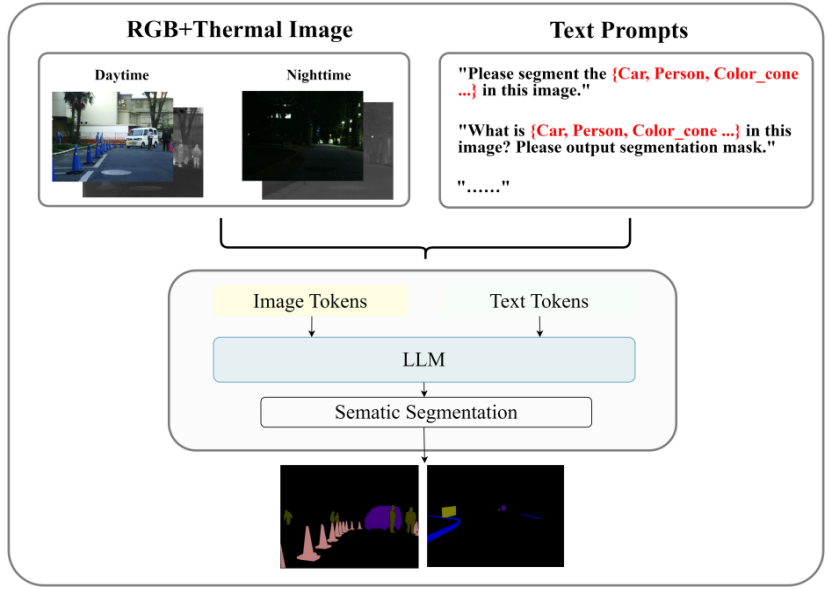

**Fig. 1.** RGB-Thermal segmentation with text prompts illustration: Employing text prompts for various scenarios, images and text tokens are fed into the LLM for processing. Leveraging the excellent perceptual abilities of the LLM, tokens containing target information are generated, facilitating RGB-Thermal semantic segmentation.

This necessitates a redesign of fusion modules for different modal data, resulting in poor generalizability. Moreover, the data fused is typically limited to image-like types, making it almost impossible to integrate textual data for the fusion process.

With the successful applications of large language models(LLMs), large multimodal models (LMMs) have emerged [14]. LMMs have powerful fitting capabilities to infer and understand information from different modalities. However, most LMMs are primarily used to process RGB and text information. According to our investigation, there are currently no fusion strategy-based LLMs for integrating RGB-Thermal data in autonomous driving scenarios, such as MFNet [1].

In this paper, we leverage the LLM to construct the **M**ultimod**A**l **S**egmentation with **TE**xt P**R**ompts (MASTER) architecture. It employs dual-path channels to process RGB and thermal images and utilizes LLM for multi-target pixel-level semantic segmentation. It can effectively fuse RGB and thermal modalities based on textual prompts, enabling multimodal data fusion of text, thermal, and RGB to generate high-quality target masks.

In summary, our contributions are as follows:
- We present the MASTER, the first work to incorporate LLM into RGB-Thermal modality fusion, enabling pixel-level semantic segmentation with textual descriptions as guidance.

- We propose a novel processing paradigm for cross-modal information extraction and specifically design a dual-path structure of Vision Transformer(ViT) for fusing RGB and thermal modalities.
- Our proposed MASTER method achieves state-of-the-art (SOTA) performance on the MFNet benchmark, a dataset specifically designed for RGB-T modality fusion tasks.

## 2 Related Work

### 2.1 RGB-Thermal Semantic Segmentation

To mitigate the impact of extreme conditions on environmental perception, researchers have made RGB-Thermal datasets and proposed multimodal semantic segmentation networks. Currently, most research focuses on designing feature fusion modules. MFNet [1] is the first RGB-T semantic segmentation model and utilizes simple convolutional pooling operations to achieve feature fusion. Some similar works [2] also employ basic channel operations for cross-modal feature fusion. ABMDRNet [3] performs fusion in a multi-scale manner through separately designed modules. CCFFNet [4] complements RGB and thermal data through channel weighting and spatial weighting fusion. CMX [5] and HAPNet [6] introduce attention mechanisms into the fusion process to enhance global information interaction. The CMNeXt [7] structure builds upon CMX by integrating non-essential modalities at the patch level, increasing the variety of fusion modalities. Sigma [8] utilizes the mamba structure for cross-scale data fusion. Some other works [9, 10] focus on enhancing network information mining capabilities through the use of random masks or generating pseudo-modalities.

Howerver, multi-scale fusion inevitably complicate model structures. Moreover, previous fusion structures often require modifications tailored to different modal data, necessitating separate modules for each, leading to poor structural generality. And leveraging data modalities with high information density like textual information becomes challenging under such circumstances.

### 2.2 Large Multimodal Models

Large Multimodal Models (LMMs) have significantly improved performance in tasks that require understanding multiple modalities. The performance of LMMs is largely determined by the capabilities of the LLMs. The emergence of LLMs has driven new directions in the development of LMMs, enhancing their multimodal understanding capabilities. Common methods within this framework include integrating adapters to align visual and textual representations within LLMs. Some notable examples include BLIP-2 [11], Flamingo [12], MiniGPT-4 [13], LLaVA [14], InstructBLIP [15], InternGPT [16], and QwenVL [17].

Existing models are primarily utilized for visual language tasks concerning visible light, and there is no architecture based on LLMs for the fusion of RGB and Thermal modalities introduced as of now.

## 3 Method

### 3.1 Overall Architecture

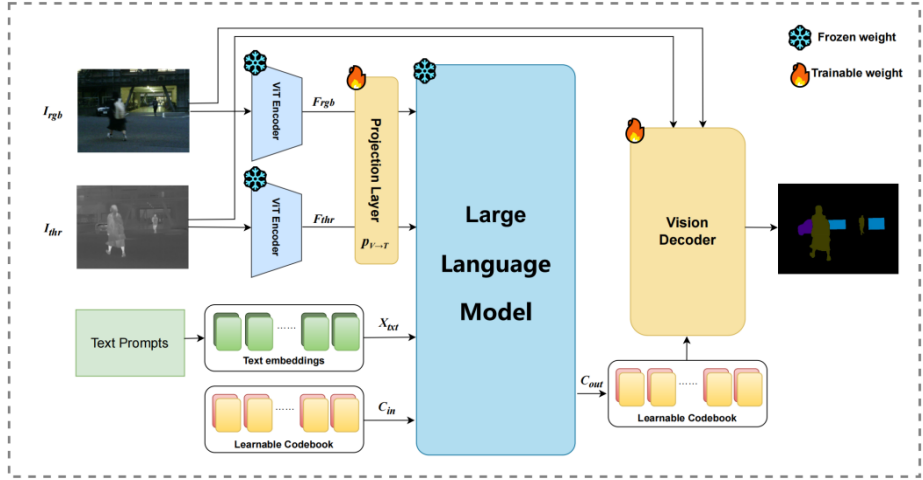

**Fig. 2.** The overall framework of our proposed MASTER. The proposed model consists of a ViT encoder for extracting image-modal features, a large multimodal language model for cross-modal feature fusion, and a lightweight pixel decoder for generating segmentation masks.

The overall framework of the MASTER is illustrated in Fig. 2. The proposed model consists of a ViT encoder for extracting image-modal features, a multimodal large language model for cross-modal feature fusion, and a lightweight pixel decoder for generating segmentation masks. This represents a pioneering approach that integrates the LLM into the fusion of RGB and thermal modalities. Specifically, the MASTER is designed around the large language model, with a visual backbone consisting of a dual-path visual encoder structure tailored for the fusion of RGB and thermal modalities. The visual backbone is based on CLIP-ViT aligned with text, which can project image tokens into language feature space.

### 3.2 Fusion with LLM

Firstly, we encode the RGB image $I_{rgb}$ and thermal image $I_{thr}$ into D-dimensional tokens $F_{rgb}$ and $F_{thr}$ of patch length using a specially designed dual-path structure encoder. Next, a projection layer maps the image features into the language space, achieving alignment in feature space. Meanwhile, the text prompts are tokenized into embeddings. The aligned image features and tokenized text information are then fed into the LLM to learn codebook tokens $C_o$, which contains information about the image targets. The large language model, being the core component of data fusion, is pretrained on a large amount of text data and exhibits excellent recognition capabilities for unseen class samples. Thus, merging different modal data using the LLM can yield high-quality tokens with target semantic information. The above processing steps are:

$$C_{out} = M(p_{V \to T}(F_{rgb}), p_{V \to T}(F_{thr}), X_{txt}, C_{in}) \tag{1}$$

where $p_{V \to T}$ presents vision-to-language projection layer, $C_{in}$ is initialized codebook tokens, $X_{txt}$ is tokens from text prompts and $M$ is LLM.

After obtaining the codebook with semantic segmentation information, we employ a decoder structure from Pixellm [22], which is similar to SAM [18] for semantic segmentation. In this process, the fused $C_{out}$ assists the mask decoder in obtaining high-quality segmentation masks.

## 4 Experiment

In this paper, we extensively evaluate the performance of the proposed model through a series of experiments. We provide a detailed overview of the dataset, experimental setups, and evaluation metrics.

### 4.1 Dataset

MFNet dataset [1] consists of 820 daytime and 749 nighttime RGB-Thermal image pairs, with a resolution of 640 × 480. This dataset provides semantic labels for 9 classes, including 1 background class and 8 common object classes: background, car, person, bike, curve, stop, guardrail, cone, and bump.

### 4.2 Implementation details

In our experiment setting, the ViT encoder was based on the fixed CLIP-ViT-L-patch14-336 model. We utilize the model LlaVA-7B and perform fine-tuning with LoRA. Before the LLM, $C_{in}$ presents the codebook, which is initialized as learnable codebook tokens. After the LLM, $C_{out}$ presents the codebook, which possesses target information. The trainable components of our model include the segmentation decoder, LoRA parameters, segmentation codebook $C_{seg}$, as well as the projection layers $p_{V \to T}$ and $p_{V \to D}$ for space mappings. Training uses a single RTX 6000 GPU with a memory capacity of 46GB, and the batch size per device is set to 2.

### 4.3 Evaluation Metrics

We employ the Intersection over Union (IoU) to quantitatively analyze the semantic understanding performance for each class. Additionally, we calculate the average values of IoU across all classes (referred to as mIoU) to represent the overall performance of the model. It can be computed as:

$$\text{mIoU} = \frac{1}{N} \sum_{i=1}^{N} \frac{\sum_{m=1}^{M} \eta_{ii}^m}{\sum_{m=1}^{M} \eta_{ii}^m + \sum_{m=1}^{M} \sum_{j=1, j \neq i}^{N} \eta_{ji}^m + \sum_{m=1}^{M} \sum_{j=1, j \neq i}^{N} \eta_{ij}^m} \tag{2}$$

where $\eta_{ij}^m$ represent the number of pixels classified as the i-th class in the j-th category in the m-th image. $M$ denotes the total number of images, and $N$ represents the total number of categories.

### 4.4 Results on MFNet

**Table 1.** Quantitative comparisons (%) with the methods on the MFNet test set. The numbers **in bold** represent the top three scores in each category.

| Method | Publication | Car | Person | Bike | Curve | Stop | Guardrail | Cone | Bump | mIoU |
|---|---|---|---|---|---|---|---|---|---|---|
| MFNet [1] | IROS 2017 | 65.9 | 58.9 | 42.9 | 29.9 | 9.9 | 0.0 | 25.2 | 27.7 | 39.7 |
| RTFNet [19] | RAL 2019 | 87.4 | 70.3 | 62.7 | **45.3** | 29.8 | 0.0 | 29.1 | **55.7** | 53.2 |
| FuseSeg [20] | TASE 2020 | **87.9** | **71.7** | **64.6** | 44.8 | 22.7 | 6.4 | 46.9 | 47.9 | 54.5 |
| ABMDRNet [3] | CVPR 2021 | 84.8 | 69.6 | 60.3 | 45.1 | 33.1 | 5.1 | 47.4 | 50.0 | 54.8 |
| FEANet [21] | IROS 2021 | 87.8 | 71.1 | 61.1 | **46.5** | 22.1 | 6.6 | **55.3** | 48.9 | 55.3 |
| CMX [5] | TITS 2023 | **90.1** | **75.2** | 64.5 | **64.5** | **50.2** | **35.3** | 8.5 | 54.2 | 60.6 |
| CRM-RGBTSeg [7] | CVPR 2023 | **90.0** | **75.1** | **67.0** | 45.2 | **49.7** | 18.4 | **54.2** | **54.4** | 61.4 |
| MASTER (Ours) | - | 86.9 | 59.4 | **66.4** | 44.1 | **47.1** | **49.4** | **53.6** | **57.8** | 62.5 |

In Table 1, we compared MASTER with other methods using the MFNet dataset [1]. Our approach simplifies modality-fusion by using projection layers to map features and leveraging LLM, achieving a significant mIoU advantage of 62.5%. Notably, MASTER excels in small objects such as guardrails and bumps, with top scores in Guardrail (+14.07% IoU) and Bump (+3.36% IoU).

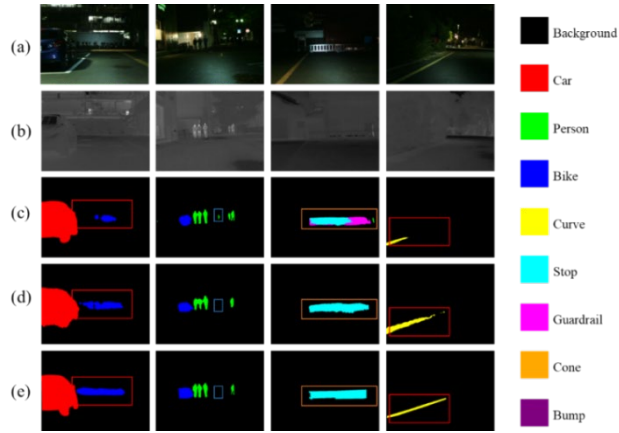

**Fig. 3.** Visual comparisons of nighttime scenes. (a) RGB images. (b) Thermal images. (c) CMNeXt [7]. (d) Ours. (e) Ground truth.

### 4.5 Qualitative Analysis

**Visualization of segmentation results.** We visualize and compare the segmentation results of the RGB-T method based on CMNeXt [7] with our method. analyzing each predicted picture in the Fig. 3 and Fig. 4, it can be observed that MASTER has the following advantages.

- Better object segmentation continuity. In our method, the objects within the red bounding boxes are more accurate compared to CMNeXt [7]. For instance, in Fig. 3 and Fig. 4, our method captures the shape of bikes and curves with more continuous segmentation results.
- Better perception of small objects. CMNeXt [7] struggles with segmenting small objects like curve and bump indicated by white boxes of Fig. 4, and our segmentation results are much clearer and better aligned with shape and position.
- Reduced noise and artifacts. In dark or challenging lighting conditions, CMNeXt [7] shows incorrect segmentation of car stops like orange boxes show in Fig. 3 and false positive results like blue boxes show in Fig. 3 and Fig. 4, while our method reduces these artifacts and shows cleaner segmentation outputs.

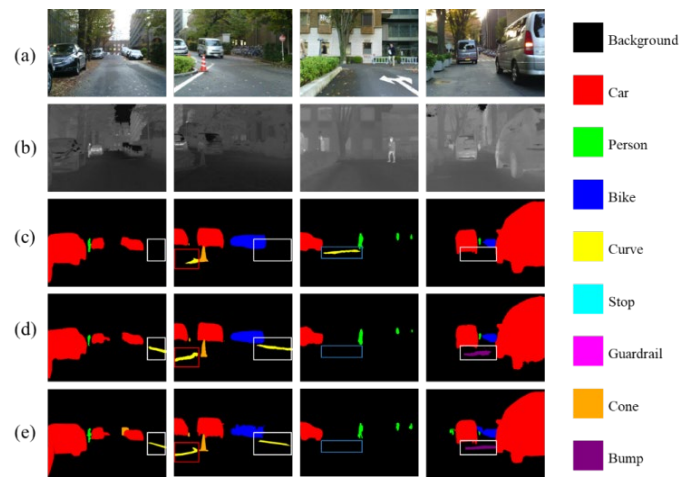

**Fig. 4.** Visual comparisons of daytime scenes. (a) RGB images. (b) Thermal images. (c) CMNeXt [7]. (d) Ours. (e) Ground truth.

## 5   Conclusion

In this study, we proposed the MASTER architecture, a multimodal fusion framework that incorporates textual cues into RGB-Thermal modal data for pixel-level image semantic segmentation. We found that it produces high-quality image masks. Moreover, through extensive comparative experiments, we achieved excellent results, demonstrating the superiority of the approach. In the future, we plan to conduct experiments in more driving scenarios.

**Acknowledgements** This work was supported by the National Natural Science Foundation of China under Grant No.U23B2034, No.62203424, and No.62176250: and the Innovation Program of the Institute of Computing Technology, Chinese Academy of Sciences under Grant No. 2024000112.